\def\year{2018}\relax
\documentclass[letterpaper]{article} %DO NOT CHANGE THIS
\usepackage{aaai18}  %Required
\usepackage{times}  %Required
\usepackage{helvet}  %Required
\usepackage{courier}  %Required
\usepackage{url}  %Required
\usepackage{graphicx}  %Required
\usepackage{graphics}
\usepackage{amsmath}
\usepackage{amssymb}
\usepackage{algorithm}
\usepackage{algpseudocode}
\usepackage{caption}		% for tabular caption
\usepackage{subfig}
\usepackage{multirow}

\begin{document}
 
\title{Regularizing Deep Networks Using Efficient Layerwise Adversarial Training}
\author{
  Swami Sankaranarayanan \\
  University of Maryland\\
  College Park, MD\\
  \texttt{swamiviv@umiacs.umd.edu} \\
  % examples of more authors
  \And
  Arpit Jain \\
  GE Global Research\\
  Niskayuna, NY\\  
  \texttt{arpit.jain@ge.com} \\
  \AND
  Rama Chellappa \\
  University of Maryland\\
  College Park, MD\\
  \texttt{rama@umiacs.umd.edu} \\
  \And
  Ser Nam Lim \\
  Avitas Systems,
  GE Global Research\\
  Boston, MA\\  
  \texttt{limser@ge.com} \\
}
  
\maketitle
\begin{abstract}

Adversarial training has been shown to regularize deep neural networks in addition to increasing their robustness to adversarial examples. However, the regularization effect on very deep state of the art networks has not been fully investigated. In this paper, we present a novel approach to regularize deep neural networks by perturbing intermediate layer activations in an efficient manner. We use these perturbations to train very deep models such as ResNets and WideResNets and show improvement in performance across datasets of different sizes such as CIFAR-10, CIFAR-100 and ImageNet. Our ablative experiments show that the proposed approach not only provides stronger regularization compared to Dropout but also improves adversarial robustness comparable to traditional adversarial training approaches.
\end{abstract}

% --------------------------------
%INTRODUCTION

\section{Introduction}\label{sec:intro}
Deep neural networks (DNNs) have shown tremendous success in several computer vision tasks in recent years [\cite{resnet},\cite{schroff2015facenet},\cite{krizhevsky2012imagenet}]. However, seminal works on adversarial examples [\cite{goodfellow2014explaining}, ~\cite{szegedy2013intriguing}] have shown that DNNs are susceptible to imperceptible perturbations at input and intermediate layer activations. From an optimization perspective, they also showed that adversarial training can be used as a regularization approach while training deep networks.  The focus of adversarial training techniques has been to improve network robustness to gradient based adversarial perturbations. In this work, we propose a novel variant of adversarial training with a focus to improve regularization performance on test data.

The proposed approach is efficient and simple to implement. It uses adversarial perturbations of intermediate layer activations to provide a stronger regularization compared to traditional techniques like Dropout (\citeauthor{srivastava2014dropout} \citeyear{srivastava2014dropout}). By generating the perturbations from a different class compared to the current input, we ensure that the resulting perturbations in the intermediate layers are directed towards an adversarial class. This forces the network to learn robust representations at each layer resulting in improved discriminability. Even though these perturbations are not adversarial at the input layer, we show that they are strongly adversarial when applied at the intermediate layers. 

The proposed regularization approach does not add any significant overhead during training and thus can be easily extended to very deep neural networks, as shown in our experiments. Our approach complements dropout and achieves regularization beyond dropout. It avoids over-fitting and generalizes well by achieving significant improvement in performance on the test set. We show that the trained network develops robustness against adversarial examples even when it is not explicitly trained with adversarial inputs. While previous works have focused on generating adversarial perturbations for standalone images, our work focuses on using these to efficiently regularize training. We perform several ablative experiments to highlight the properties of the proposed approach and present results for very deep networks such as VGG~\cite{vgg}, ResNets~\cite{resnet} and state of the art models such as WideResNets \cite{wideresnet} on CIFAR-10, CIFAR-100 and ImageNet datasets.

% --------------------------------
% RELATED WORKS
\section{Related Work}

Many approaches have been proposed to regularize the training procedure of very deep networks. Early stopping and statistical techniques like weight decay are commonly used to prevent overfitting. Specialized techniques such as DropConnect ~\cite{icml2013_wan13}, Dropout ~\cite{srivastava2014dropout} have been successfully applied with very deep networks. Faster convergence of such deep architectures was made possible by Batch Normalization (BN) \cite{ioffe2015batch}. One of the added benefit of BN was that the additional regularization provided during training even made dropout regularization unnecessary in some cases.

The work of Szegedy \emph{et al.} ~\cite{szegedy2013intriguing} showed the existence of adversarial perturbations for computer vision tasks by solving a box-constrained optimization approach to generate these perturbations. They also showed that training the network by feeding back these adversarial examples regularizes the training and makes network resistant to adversarial examples. Due to a relatively expensive training procedure, their analysis was limited to small datasets and shallow networks.~\cite{goodfellow2014explaining} proposed the fast gradient sign method to generate such adversarial examples. They proposed a modified loss function with an adversarial objective that improves network robustness. ~\cite{miyato2017virtual} proposed a virtual adversarial training framework and showed its regularization benefits for relatively deep models, while incurring additional computational overhead during training. Furthermore, recent approaches such as deep contrastive smoothing~\cite{gu2014towards}, distillation  ~\cite{papernot2016distillation} and stability training~\cite{zheng2016improving} have focused solely on improving the robustness of deep models to adversarial inputs. In this work, we present an efficient layerwise approach to adversarial training and demonstrate its ability as a strong regularizer for very deep models beyond the specialized methods mentioned above, in addition to improving model robustness to adversarial inputs.

Recent theoretical works [\cite{fawzi2015analysis}, ~\cite{fawzi2016robustness}] analyze the effect of random, semi-random and adversarial perturbations on classifier robustness. They presented fundamental upper bounds on the robustness of classifier which depends on factors such as curvature of decision boundary and distinguishability between class cluster centers. Wang \emph{et al.} \cite{WangGQ16} point out that the differences between generalization and robustness by characterizing the topology of the learned classification function. In this work, we perform empirical studies that show that the proposed approach improves performance by suppressing those dimensions that are unnecessary for generalization. In addition, we observe that pure adversarial training techniques suppress most dimensions resulting in strong robustness against adversarial examples but less improvement in generalization. To the best of our knowledge, this is the first work that provides a comparison between regularization and adversarial robustness by providing empirical results on very deep networks.

% ---------------------------------
% METHODS

\section{Our Approach}\label{sec:method}
In this section, we present the proposed regularization approach and highlight the  differences between related methods that use adversarial training for regularization. In addition, we perform a small scale experiment to study the properties of the proposed approach by analyzing the singular value spectrum of the Jacobian. We also visualize the impact of these perturbations on the intermediate layer activations and conclude by illustrating the connection to robust optimization.

%\subsection*{Notation}
We start by defining some notation. Let $\{x_i\}_{i=1}^{N}$ denote the set of images and $\{y_i\}_{i=1}^{N}$ denote the set of labels. Let $f: x \in \mathbb{R}^{m} \mapsto y \in \mathbb{L}$ denote the classifier mapping that maps the image to a discrete label set, $\mathbb{L}$. In this work, $f$ is modeled by a deep CNN unless specified otherwise. We denote the loss function of the deep network by $\mathcal{J}(\theta,x,y)$ where $\theta$ represents the network parameters and $\{x,y\}$ are the input and output respectively. The deep network consists of $L$ layers and $\nabla_{l}\mathcal{J}(\theta^{t},x^{t},y^{t})$ denotes the backpropagated gradient of the loss function at the output of the ${l}^{th}$ layer at iteration $t$. In the above expression, $l=0$ corresponds to the input layer and $l=L-1$, the loss layer. Let $x^{t}_{l}$ be the input activation to the $l^{th}$ layer and $r^{t}_{l}$ represents the perturbation that is added to $x^{t}_{l}$. For clarity, we drop the subscript $l$ when talking about the input layer. %Update this as required...

\subsection{Overview of Adversarial training methods}
Previous works on adversarial training have observed that training the model with adversarial examples acts as a regularizer and improves the performance of the base network on the test data. \citeauthor{szegedy2013intriguing} \shortcite{szegedy2013intriguing} define adversarial perturbations $r$ as a solution of a box-constrained optimization as follows: Given an input $x$ and target label $y$, they intend to minimize $||r||_{2}$ subject to (1) $f(x+r)=y$ and (2) $x+r \in [0,1]^{m}$. Note that, if $f(x)=y$, then the optimization is trivial (i.e. $r=0$), hence $f(x)\neq y$.  While the exact minimizer is not unique, they approximate it using a box-constrained L-BFGS. More concretely, the value of $c$ is found using line-search for which the minimizer of the following problem satisfies $f(x+r)=\hat{y}$, where $\hat{y}\neq y$:
\begin{equation}
\underset{r}{\text{argmin}} \;  c||r||_{2} + \mathcal{J}(\theta,x+r,y), \quad \text{subject to} \quad x + r \in [0,1]^{m}
\label{eq:lbfgs}
\end{equation}
This can be interpreted as finding a perturbed image $x+r$ that is closest to $x$ and is misclassified by $f$. The training procedure for the above framework involves optimizing each layer by using a pool of adversarial examples generated from previous layers. As a training procedure, this is rather cumbersome even when applied to shallow networks having 5-10 layers. To overcome the computational overhead due to the L-BFGS optimization performed at each intermediate layer, \cite{goodfellow2014explaining} propose the Fast Gradient Sign (FGS) method to generate adversarial examples. By linearizing the cost function around the value of the model parameters at a given iteration, they obtain a norm constrained perturbation as follows: $r_{fgs}=\epsilon . sign(\nabla\mathcal{J}(\theta,x,y))$. They show that the perturbed images $x+r_{fgs}$ reliably cause deep models to misclassify their inputs. As noted in \cite{understandingAT}, the above formulation for adversarial perturbation can be understood by looking at a first order approximation of the loss function $\mathcal{J}(\cdot)$ in the neighborhood of the training sample $x$:

\begin{equation}
\tilde{\mathcal{J}}(\theta,x+r,y) = \mathcal{J}(\theta,x,y)+ \langle \nabla\mathcal{J}(\theta,x,y),r\rangle
\end{equation}

The FGS solution ($r_{fgs}$) is the result of maximizing the second term with respect to $r$, with a $l_{\infty}$ norm constraint. They train the network with the following objective function with an added adversarial objective:
\begin{equation}
\tilde{\mathcal{J}}(\theta,x,y) = \alpha \mathcal{J}(\theta,x,y)+ (1-\alpha) \mathcal{J}(\theta,x+r_{fgs},y)
\label{eq:fgs_train}
\end{equation}

By training the model with both original inputs and adversarially perturbed inputs, the objective function in \ref{eq:fgs_train} makes the model more robust to adversaries and provides marginal improvement in performance on the original test data. Intuitively, the FGS procedure can be understood as perturbing each training sample within a $L_{\infty}$ ball of radius $\epsilon$, in the direction that maximally increases the classification loss. 

The focus of the adversarial training techniques described above is to improve a model's robustness to adversarial examples. As a by-product, they observe that an adversarial loss term can also marginally regularize the deep network training. In this work, we propose a novel approach which is an extension of the traditional adversarial training. The focus of our approach is to improve regularization and as an interesting by-product, we observe improvement in adversarial robustness of the trained model as well.

\subsection{Proposed Formulation}\label{subsec:formulation}
The proposed approach differs in the following aspects compared to the formulations discussed above: (1) Generating adversarial perturbations from intermediate layers rather than just using the input layer (2) Using the gradients from the previous batch to generate adversarial perturbations for the activations of the current batch. In order to facilitate the representation of the intermediate activations in the loss function, we denote the collection of layerwise responses as $X=\{x_l\}_{l=0}^{L-1}$ and  the set of layerwise perturbations as $R=\{r_l\}_{l=0}^{L-1}$. Then, $\mathcal{J}(\theta,X+R,y)$ denotes the loss function where intermediate layer activations are perturbed according to the set $R$. The notation used in the previous section is a special case where $X={x}$ and $R=r_{fgs}$. Now, consider the following objective to obtain the perturbation set $R$: 

\begin{equation}
\begin{aligned}
& \underset{R}{\text{argmax}} \; \mathcal{J}(\theta,X+R,y) \\
& \text{subject to} \quad ||r_{l}||_{\infty} \leq \epsilon, \; \forall l, \; f(X+R) \neq y
\label{eq:our_train}
\end{aligned}
\end{equation}

Ideally, for each training example $x$, the solution to the above problem, consists of generating the perturbation corresponding to the maximally confusing class; in other words, by choosing the class $\hat{y}$ which maximizes the divergence, $KL(p(y|x_{L-1}),p(\hat{y}|x_{L-1}))$.  In the absence of any prior knowledge about class cooccurences, solving this explicitly for each training sample for every iteration is time consuming. Hence we propose an approximate solution to this problem: the gradients computed from the previous sample at each intermediate layer are cached and used to perturb the activations of the current sample. In a mini-batch setting, this amounts to caching the gradients of the previous mini-batch. To ensure that the class constraint in Eq. \ref{eq:our_train} is satisfied, the only requirement is that successive batches have little lateral overlap in terms of class labels. From our experiments, we observed that any random shuffle of the data satisfies this requirement. For more discussion on this, please refer to the experiments section. Given this procedure of accumulating gradients, we are no longer required to perform an extra gradient descent-ascent step as in the FGS method to generate perturbations for the current batch. Since the gradient accumulation procedure does not add to the computational cost during training, this can be seamlessly integrated into any existing supervised training procedure including even very deep networks as shown in the experiments. 

\begin{algorithm*}[!htb]
\caption{Efficient layerwise adversarial training procedure for improved regularization}\label{alg:algo}
\label{tab:algo}
\begin{algorithmic}[1]
\State Inputs: Deep network $f$ with loss function $\mathcal{J}$ and parameters $\theta$ containing C convolutional blocks. ${B^t}$ is the batch sampled at iteration $t$ of size $k$, with input-output pairs $\{X^{t},Y^{t}\}$. Gradient accumulation layers $\{G_c\}_{c=1}^{C}$, with stored perturbations $R^t=\{r^{t}_c\}_{c=1}^{C}$, initialized with zero. Perturbation parameter, $\epsilon$.
\State {t=0:}
		\State Sample a batch $\{X^t,Y^t\}$ of size $k$ images from the training data
        \State Perform regular forward pass - $G_c$'s are not active for $t=0$.
        \State Perform backward pass using the classification loss function. Each gradient accumulation layer $G_{c}$ stores the gradient signal backpropagated to that layer:
        \begin{equation}
        r^{t+1}_c = sign(\nabla_{c}\mathcal{J}(\theta^t,X^t+R^t,Y^t)),  \forall c = [1,C]
        \label{eq:update}
        \end{equation}
\For{t in 1:$|B|-1$}
		
		\State Sample a batch $\{X^t,Y^t\}$ of size $k$ from the training data
        \State Perform forward pass with perturbation: Each gradient accumulation layer acts as follows. \noindent Let $X^{t}_{c}$ be the input to block $c$, then:
        \begin{equation}
        		G_c(X^{t}_{c})=X^{t}_{c}+\epsilon \cdot r^{t}_c
         \end{equation}
        \State Perform backward pass updating $r^{t}_c$ to $r^{t+1}_c$ for all blocks $c$ as in Eq. ~\ref{eq:update} above.
\EndFor
\State \textbf{Test time usage:} During test time, the gradient accumulation layers ($G_c$) are removed and $f$ behaves as a standard feed forward deep network.
\end{algorithmic}
\end{algorithm*}

 The training procedure is summarized in Algorithm \ref{tab:algo}, where $sign(\cdot)$ denotes the signum function. We add the gradient accumulation layers after the Batch Normalization (BN) layer in each convolutional block (conv-BN-relu). In case the BN layers are not present, we add gradient accumulation layers after each convolution layer. A subtle detail that is overlooked in the algorithm is that the value of $\epsilon$ is not constant over all the layers, rather it is normalized by multiplying with the range of the gradients generated at the respective layers. During test time, the gradient accumulation layers ($G_c$) are removed from the trained model.

\begin{figure*}[!th]
  \centering
  \subfloat[$\epsilon=0, 91.9\% $]{\includegraphics[scale=0.23]{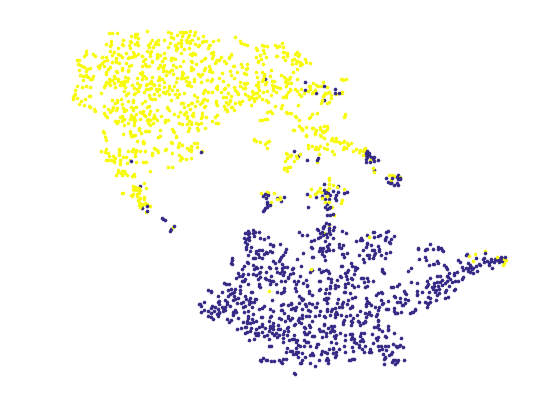}\label{fig:f1}}
  \hfill
  \subfloat[$\epsilon=0.02, 69.4\%$]{\includegraphics[scale=0.23]{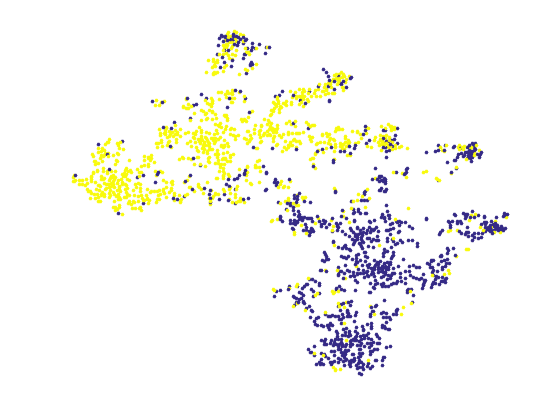}\label{fig:f2}}
  \hfill
  \subfloat[$\epsilon=0.04, 48.2\%$]{\includegraphics[scale=0.23]{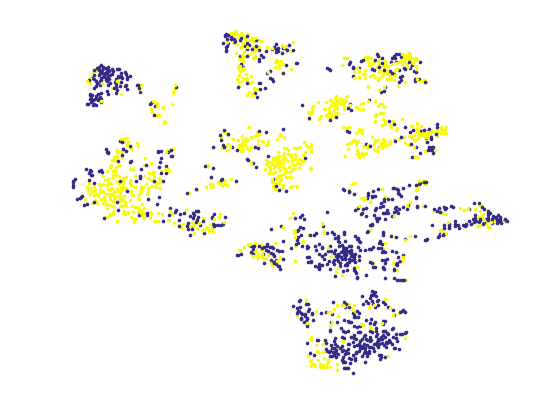}\label{fig:f3}}
  \hfill
  \subfloat[$\epsilon=0.06, 34.6\%$]{\includegraphics[scale=0.23]{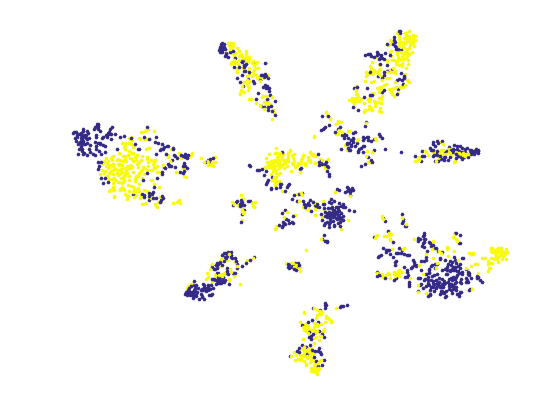}\label{fig:f4}} \\
   
  \subfloat[$\epsilon=0, 91.9\%$]{\includegraphics[scale=0.23]{eps_0_0.png}\label{fig:f1}}
  \hfill
  \subfloat[$\epsilon=0.02, 91.85\%$]{\includegraphics[scale=0.23]{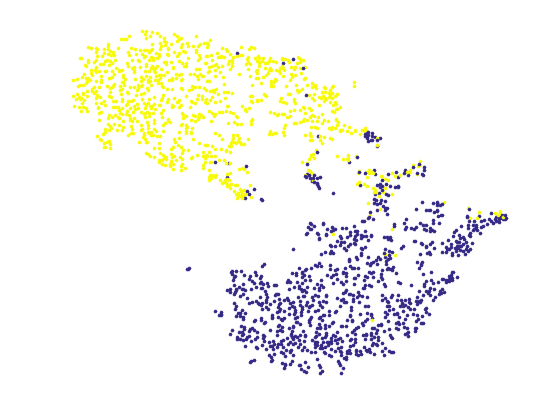}\label{fig:f2}}
  \hfill
  \subfloat[$\epsilon=0.04, 91.9\%$]{\includegraphics[scale=0.23]{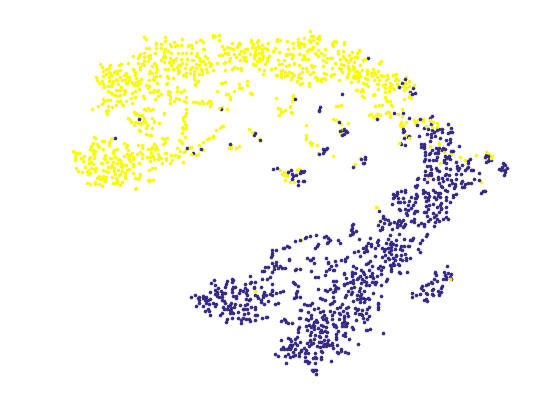}\label{fig:f3}}
  \hfill
  \subfloat[$\epsilon=0.06, 85.8\%$]{\includegraphics[scale=0.23]{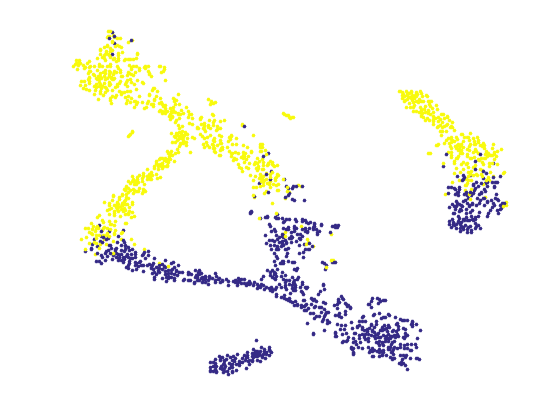}\label{fig:f4}}
  \caption{t-SNE visualization of the final fc-layer features of dimension 512 of the VGG network for two randomly chosen classes of the CIFAR-10 data for different values of the intensity, $\epsilon$. The top row shows the effect of the perturbations generated using the proposed approach while the bottom row shows random perturbations of the same intensity. It is clear that the random perturbations do not affect the linear separability of the data, while the proposed perturbations are extremely effective in leading the network to misclassify the perturbed data.}
\label{fig:adv_tsne}
\end{figure*}

To understand the effect of the proposed layerwise perturbations described in the previous section, we compare them with random layerwise perturbations. Figure \ref{fig:adv_tsne} shows a two dimensional t-SNE \cite{maaten2008visualizing} visualization of the embeddings belonging to the penultimate FC-layer for a range of values of $\epsilon$, the intensity of the adversarial perturbation. 

We used a pretrained VGG network that was trained on the CIFAR-10 dataset to compute the embeddings for two randomly chosen classes from the test data. In the bottom row, the effect of random perturbations with zero mean and unit standard deviation, applied layerwise on the original data is also shown. From the visualization and the accuracy values, it is clear that the proposed perturbations when added to the original data makes the network misclassify the original data. Hence training using these perturbations results in good regularization and improved robustness. Notice that even for higher values of $\epsilon$, the data perturbed by layerwise random gradient directions remains clearly linearly separable while the data perturbed by the accumulated gradients is unable to be distinguished by the base model. 

\subsection{Toy example}\label{subsec:toy-eg}
In order to acquire a better understanding of the regularizing properties of the mapping function learned using the proposed approach, we perform a toy experiment using a small neural network consisting of two fully connected layers of sizes 1024 and 512. Each fully connected layer is followed by a hyperbolic tangent activations. We use the grayscale version of the CIFAR-10 dataset as our testbed and $L_2$ norm weight decay was applied during training. No data augmentation or other regularization methods such as dropout were used during training. We train three networks: a baseline network, a network with the gradient accumulation layers (as in Algorithm 1) added after each fully connected layer and a network using the FGS training approach. Cross entropy loss was used to train all the networks. In terms of classification accuracy, the proposed method improves the baseline performance from 39.5\% to 43.3\% on the original data while the accuracy of the FGS network is 40.5\%. 

\begin{figure}[h!]
\centering
\includegraphics[scale=0.3]{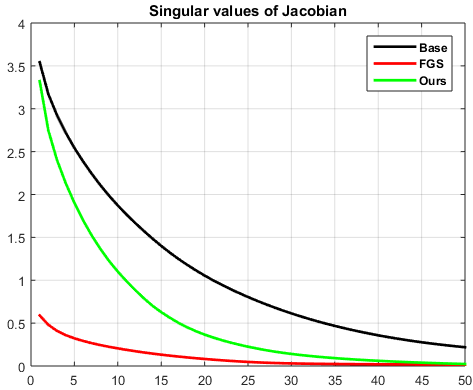}
\caption{Average singular value (SV) spectrum showing top 50 SVs for the toy example presented in the text. A model regularized with the proposed approach is compared with a FGS regularized model and baseline model with no regularization.}
\label{fig:jacobian}
\end{figure}

\textbf{Singular Value Analysis:} To gain a deeper understanding of the encoder mapping learned by each network, we perform an analysis similar to \cite{rifai2011contractive} by computing the singular values of the Jacobian of the encoder. Since this is a small architecture, we are able to explicitly compute the Jacobian for each sample in the test set. The average singular value spectrum of the Jacobian for the test data is shown in Figure \ref{fig:jacobian}. We can make the following observations: (a) The singular value spectrum computed for ours and FGS approach has fewer dominant singular values and decays at a much faster rate compared to base network (b) The FGS training suppresses the response of the network strongly in all the dimensions while our approach achieves a strong suppression only for trailing dimensions. This implies that our network is able to better capture data variations that are relevant for classifying original test data hence providing improved performance. On the other hand, FGS achieves slightly improved robustness against adversarial examples compared to our approach by suppressing the network's response strongly even in leading dimensions, as demonstrated in our ablative experiments in the next section. 

\subsection{Connection to Robust Optimization}
Several regularization problems in machine learning such as ridge regression, lasso or robust SVMs have been shown to be instances of a more general robust optimization framework \cite{sra2012optimization}. To point out the connection between the proposed approach and robust optimization, we borrow the idea of uncertainty sets from \cite{understandingAT}. To explain briefly, an uncertainty set denoted by $\mathcal{U} = \mathcal{B}_{\rho}(x,\epsilon)$ represents an epsilon ball around $x$ under norm $\rho$. \cite{goodfellow2014explaining} point out that adversarial training can be thought of as training with hard examples that strongly resist classification. 
%They also perturb the inputs using random noise by an uniform $\epsilon$ ball around each sample and report no difference in original or adversarial performance. 
Under the setting of uncertainty sets, adversarial training with the FGS method could be seen as generating the worst case perturbations from the input space $\mathcal{U}$ under the $l_{\infty}$ norm. In this work, we extend the idea of uncertainty sets from input activations to intermediate layer activations. This can be thought of as sampling perturbations from the feature space learned by the deep network. 
Let $\mathcal{U}_{l}$ represent the uncertainty set of the activation $x_{l}$ at layer $l$. Then, the proposed adversarial training approach is equivalent to sampling perturbations from the intermediate layer uncertainty sets which makes the feature representation learned at those layers to become more robust during training. Moreover, by generating perturbations from inputs that do not belong to the same class as the current input, the directions sampled from the uncertainty set tend to move the perturbed feature representation towards the direction of an adversarial class. This effect can be observed from the t-SNE visualization shown in Figure \ref{fig:adv_tsne}.

% ------------------------------------
% RESULTS

\section{Experiments}
\begin{figure*}[!th]
  \centering
  \subfloat[CIFAR-10]{\includegraphics[scale=0.2]{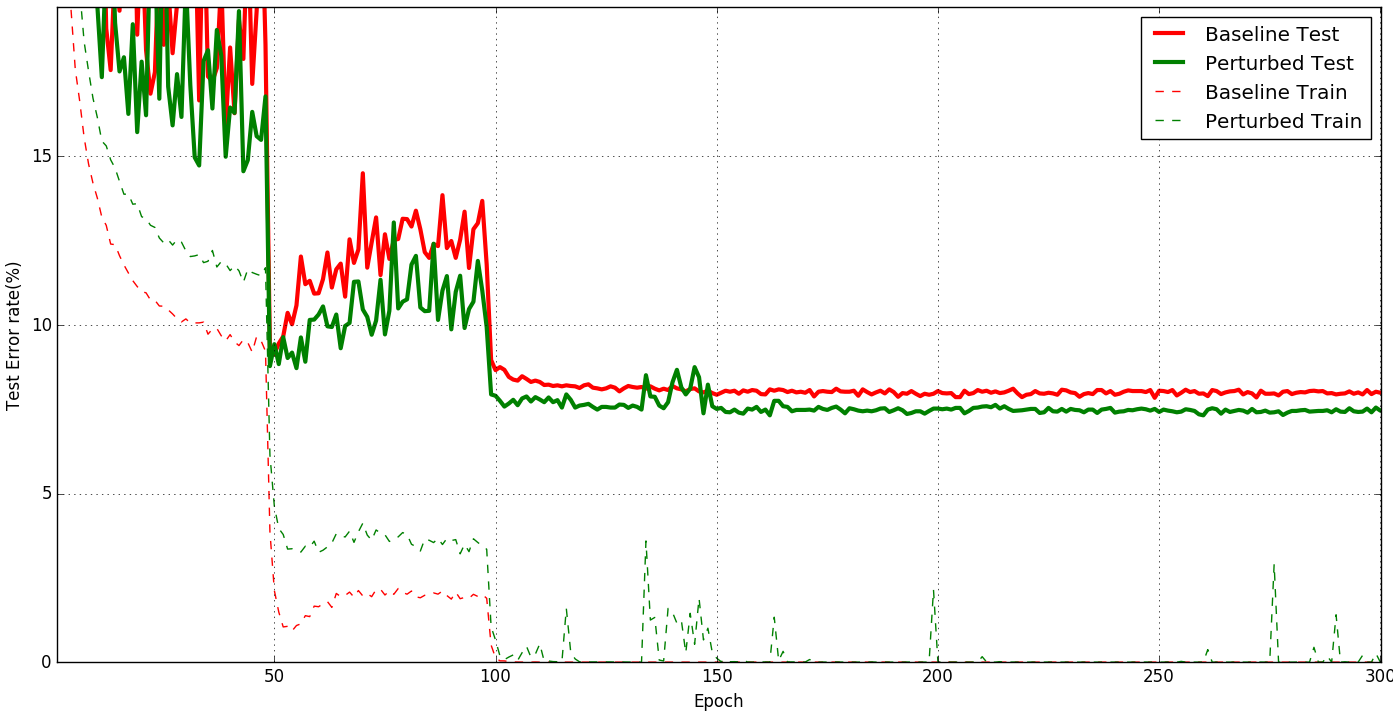}}   
   \subfloat[CIFAR-100]{\includegraphics[scale=0.2]{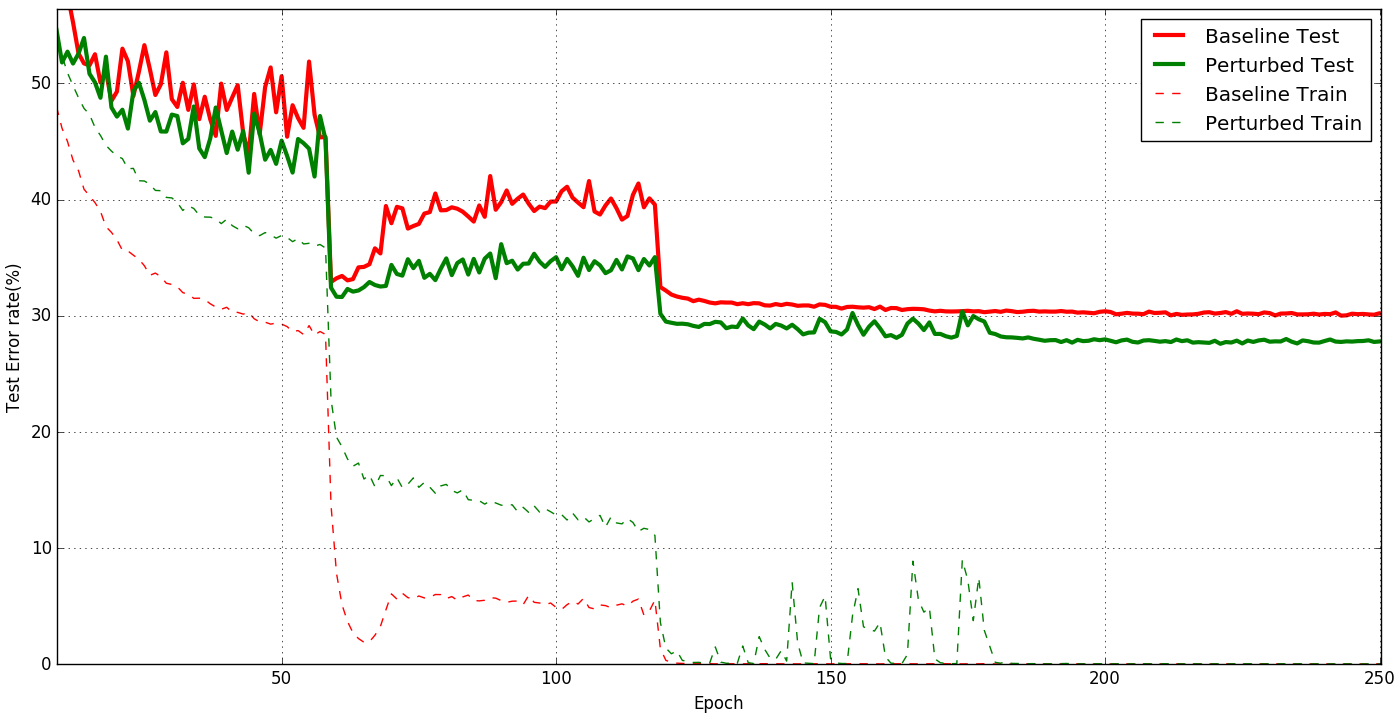}}  
  \caption{Training and test error rates for VGG network trained on CIFAR-10 and CIFAR-100 datasets. The training errors are computed using  perturbed activations in each epoch. The red color indicates the baseline model and green indicates the model regularized with the proposed approach, referred as \textit{Perturbed} in the text. Best viewed in color. Please zoom in for clarity.}
  \label{fig:train_example}
\end{figure*}

In this section, we provide an experimental analysis of the proposed approach to show that layerwise adversarial training improves the performance of the model on the original test data and increases robustness to adversarial inputs. To demonstrate the generality of our training procedure, we present results on CIFAR-10 and CIFAR-100 \cite{krizhevsky2009learning} using VGG, ResNet-20 and ResNet-56 networks. For the ResNet networks, we use the publicly available torch implementation \cite{ResNetTorch}. For the VGG architecture, we use a publicly available implementation which consists of Batch Normalization \cite{VGGTorch}. For all the experiments, we use the SGD solver with Nesterov momentum of 0.9. The base learning rate is 0.1 and it is dropped by 5 every 60 epochs in case of CIFAR-100 and every 50 epochs in case of CIFAR-10. The total training duration is 300 epochs. We employ random flipping as a data augmentation procedure and standard mean/std preprocessing was applied conforming to the original implementations. For the ResNet baseline models, without regularization, we find that they start overfitting if trained longer and hence we perform early stopping and report their best results. For the perturbed models, we find that no early stopping is necessary; the learning continues for a longer duration and shows good convergence behavior.  We refer to the model trained using the proposed approach described in Algorithm \ref{tab:algo} as $Perturbed$ throughout this section. Dropout was not used in any of the training procedure in this experiment. We explicitly compare our approach against dropout in the ablative experiments. Figure \ref{fig:train_example} plots the training and test error rates for the baseline model and the proposed approach. It can be observed that our method converges faster and achieves better generalization error. 
\begin{center}
\captionof{table}{Classification accuracy (\%) on CIFAR-10 and CIFAR-100 for VGG and Resnet architectures. Results reported are average of 5 runs.}
\label{tab:cifar-results}
 \begin{tabular}{||c | c | c ||} 
 \hline
 Type (CIFAR-10) & Baseline & Perturbed \\ [0.5ex] 
 \hline
 VGG  & 92.1 $\pm$ 0.3 & \textbf{92.65 $\pm$ 0.2} \\ 
 \hline
 Resnet-20 & 90.27 $\pm$ 0.4 & \textbf{91.1 $\pm$ 0.3} \\
 \hline  
 Resnet-56 & 91.53	$\pm$ 0.3 & \textbf{94.1 $\pm$ 0.2} \\
 \hline 
\end{tabular}
\quad 
\begin{tabular}{||c | c | c ||} 
 \hline
 Type (CIFAR-100) & Baseline & Perturbed \\ [0.5ex] 
 \hline
 VGG  & 69.8 $\pm$ 0.5 & \textbf{72.3 $\pm$ 0.3}  \\ 
 \hline
 Resnet-20 & 64.0 $\pm$ 0.2 & \textbf{66.9 $\pm$ 0.3} \\
 \hline
 Resnet-56 & 68.2 $\pm$ 0.4 & \textbf{71.4 $\pm$ 0.5} \\
 \hline
\end{tabular}
\end{center}

\textbf{Imagenet Experiment:} To test the applicability of our regularization approach over a large scale dataset, we conducted an experiment using the ImageNet dataset (train: 1.2M images, val: 50K images). We used AlexNet as the base architecture. We used the publicly available implementation from the torch platform \cite{AlexnetTorch} and both the baseline and the regularized models were trained from scratch to 60 epochs. The classification accuracies obtained were: Baseline - 56.1\%, Proposed - 59.2\%, an increase of 3.1\%. This shows that our approach can significantly improve the performance of deep neural networks even on a large and diverse corpus like Imagenet.

% \textbf{Note on data shuffling:} As mentioned in our formulation, we employ random shuffling when training our models. However, one can argue that the performance of the proposed approach can be improved by performing a controlled data shuffling at each iteration such that there is no class label overlap between successive mini-batches. We found that this did not yield statistically significant improvement in performance and hence we use only random shuffling in all our experiments. This can be explained by our observation that even with random shuffling the overlap between successive batches was $\sim$3\% for CIFAR-10 dataset and it was even lesser for the CIFAR-100 and ImageNet datasets. We would like to note here that in a setting where the training dataset has a severely unbalanced distribution, an inexpensive data shuffling strategy can be employed once before training begins or at the beginning of each epoch to ensure no overlap. 

\section{Ablative Experiments and Discussion}

\textbf{Comparison with Dropout:} We perform an experiment where we compare the regularization performance of the proposed adversarial training approach to Dropout. We use the VGG architecture used in the previous sections and perform experiments with and without dropout on CIFAR-10 and CIFAR-100 datasets. 

\begin{center}
\captionof{table}{Comparison of classification accuracy (\%) with/without dropout on CIFAR-10 and CIFAR-100 for the VGG model}
\label{tab:dropout-results}
 \begin{tabular}{||c | c | c ||} 
 \hline
 Type (CIFAR-10) & Baseline & Perturbed \\ [0.5ex] 
 \hline
 w/o Dropout & 92.1 & 92.7  \\ 
 \hline
 with Dropout & 92.5 & 93.2 \\
 \hline  
\end{tabular}
\quad 
\begin{tabular}{||c | c | c ||} 
 \hline
 Type (CIFAR-100) & Baseline & Perturbed \\ [0.5ex] 
 \hline
 w/o Dropout & 69.8 & 72.3  \\ 
 \hline
 with Dropout & 70.5 & 73.1 \\
 \hline  
\end{tabular}
\end{center}

The following observations can be made from Table \ref{tab:dropout-results}: (1) The perturbed model performs better than the baseline model with or without dropout. Thus, the proposed training improves the performance of even dropout based networks. (2) On a complex task like CIFAR-100, the proposed adversarial training based regularization gives better performance compared to that provided by dropout (70.5\% (vs) 73.1\%). Since the proposed adversarial perturbations are intended to move the inputs towards directions that strongly resist correct classification, they are able to create a more discriminative representation for tasks with a larger number of classes. 

\begin{center}
\captionof{table}{Effect of adding gradient accumulation layers incrementally (from shallow to deeper layers) throughout the deep network. The numbers reported are the classification accuracy using the VGG network on the CIFAR-10 dataset. Baseline performance is 89.4\%. \textit{conv1 to} indicates we start adding the gradient accumulation layers from \textit{conv1} upto the mentioned layers such as \textit{pool1}, \textit{pool2} etc.}
\label{tab:layerwise}
\resizebox{.45\textwidth}{!}{%
 \begin{tabular}{||c | c | c | c | c | c ||} 
 \hline
 Layer (conv1 to) & pool1 & pool2 & pool3 & pool4 & pool5\\ [0.5ex] 
 \hline
 Accuracy & 89.5 & 89.62 & 90.24 & 91.1  & 91.3  \\
 \hline  
\end{tabular}
}
\end{center}

\textbf{Perturbing deeper layers:} In this section, we analyze the effect of adversarial perturbations starting from the lowest convolutional layers which model edges/shape information to the more deeper layers which model abstract concepts. For this experiment, we use the VGG network with batch normalization that was used in the previous section. The experiments were performed on the CIFAR-10 dataset. No data augmentation or dropout is applied. It is clear from the results in Table \ref{tab:layerwise} that the improvement in performance due to the proposed layerwise perturbations become significant when applied to the deeper layers of the network, which is in line with the observation made by \cite{szegedy2013intriguing}. While performing layerwise alternate training as proposed by\cite{szegedy2013intriguing}  becomes infeasible for even moderately deep architectures, our training scheme provides an efficient framework to infuse adversarial perturbations throughout the structure of very deep models.

\begin{table}
\centering
\captionof{table}{Comparison of the strength of adversarial examples between the FGS approach applied at the input and using the proposed layerwise perturbations. Reported numbers are classification accuracies for different values of $\epsilon$.}
\label{tab:compare_adv}
\resizebox{0.45\textwidth}{!}{%
 \begin{tabular}{||c | c | c | c | c | c ||} 
 \hline
 Type & $\epsilon=0$ & $\epsilon=0.05$ & $\epsilon=0.1$ & $\epsilon=0.15$ & $\epsilon=0.2$ \\ [0.5ex] 
 \hline\hline
 FGS \shortcite{goodfellow2014explaining}  & 92.4 & 53.28 & 41.58 & 36.44 & 33.85  \\ 
 \hline
%  Ours - only input & 92.4 & 92.35 & 92.29 & 92.04 & 90.91 \\
%  \hline 
 Ours - all layers & 92.4 & 48.56 & 21.72 & 14.76 & 14.19 \\
 \hline
\end{tabular}
}
\end{table}
\textbf{Adversarial strength of the proposed perturbations:} Traditional adversarial training techniques improve the performance on adversarial examples by explicitly making the network robust to adversarial gradient directions. Thus, an important question that needs to be addressed in light of the proposed optimization strategy is: Are the gradient directions generated from the previous mini-batch (as described in Algorithm 1) adversarial? To answer this question, we perform an empirical experiment to measure the performance of a deep model (3 convolutional layers + 1 fc-layer) trained using CIFAR-10 training data, on the CIFAR-10 test data. No adversarial training was used to train this model. As described earlier, for each test sample, the intermediate layer activations are perturbed using gradients accumulated from the previous sample. For comparison, we also show the performance of the same model on the adversarial data generated using the FGS method. From the metrics in Table \ref{tab:compare_adv}, it can be observed that using accumulated gradients from the previous batch as adversarial perturbations results in a bigger drop in performance. This signifies that the aggregated effect of layerwise perturbations is more adversarial compared to perturbing only the input layer as done in the FGS approach. We performed an additional experiment where only the input layer was perturbed using the gradients of the previous sample instead of perturbing all the intermediate layers. We found that this resulted in negligible drop in the baseline performance, indicating that these gradients are not adversarial enough when used to perturb only the input. 

\textbf{Variants of the proposed approach:}\label{subsec:variants} In the proposed training method summarized in Algorithm \ref{tab:algo} (referred as Ours-orig), each batch of inputs is perturbed at intermediate layers by the gradients accumulated from the previous batch. In this section, we present an empirical comparison between the following variants:
\begin{itemize}
\item FGS-orig: The original FGS joint loss based adversarial training as proposed by \cite{goodfellow2014explaining} and shown in Eq. (\ref{eq:fgs_train}). We used a value of $\alpha=0.5$; we did not find other values yield any significant improvements. %Following the original approach, the FGS perturbations are only used at the input.
\item FGS-inter: In this setting, different from \cite{goodfellow2014explaining}, we use the FGS gradients to perturb the intermediate layer activations and use the joint loss with $\alpha=0.5$.
\item Ours-joint: This setting is same as Ours-orig with the exception that we use the joint loss formulation with $\alpha=0.5$. Note that, Ours-orig corresponds to setting where $\alpha=0$
\end{itemize}

\begin{table}[h!]
\begin{center}
\captionof{table}{Comparison of classification accuracy (\%) between the different variants of the proposed approach and FGS method (FGS-orig) for different values of $\epsilon$}
\label{tab:variants}
\resizebox{.45\textwidth}{!}{%
 \begin{tabular}{||c | c | c | c | c | c | c ||} 
 \hline
 Type & $0$ & $0.02$ & $0.04$ & $0.06$ & $0.08$ & $0.1$ \\
 \hline
 Baseline & 89.4 & 67.5 & 49.6 & 41.2 & 37.3 & 34.7   \\
 \hline  
 FGS-orig \shortcite{goodfellow2014explaining} & 88.7 & 86.4 & 84.1 & 81.4 & 80.5 & 77.1  \\
 \hline  
 FGS-inter & 90.9 & 87.79 & 83.85 & 79.65 & 74.69 & 69.92 \\
 \hline  
 Ours-orig & 91.2 & 87.95 & 83.84 & 79.11  & 73.66 & 68.37  \\
 \hline  
 Ours-joint & 91.5 & 86.07 & 81.38 & 75.72 & 70.25 & 64.76  \\
 \hline  
\end{tabular}%
}
\end{center}
\end{table}

All the models are trained on the CIFAR-10 dataset. No data augmentation or dropout regularization is applied. The training parameters are similar to the ones used in the previous section. We generate adversarial test data for the CIFAR-10 test dataset using the FGS method, since it has been shown to generate adversarial examples reliably. We then test the models on the original and adversarial test data for different values of the adversarial strength $\epsilon$. Table \ref{tab:variants} shows the results of the different training strategies. $\epsilon=0$ corresponds to the original test data and other values of $\epsilon$ indicate the strength of adversarial FGS perturbation added to the input image. From these results, we make the following observations: (1)  Approaches based on perturbing intermediate layers (FGS-inter,Ours-orig,Ours-joint) improve the performance on the original data significantly as compared to perturbing only the input but they marginally decrease the adversarial test performance. (2) On the other hand, perturbing only the input layer (FGS-orig) yields the best adversarial test performance among the compared approaches while performing marginally worse than the baseline on the original test data. These observations indicate the possibility of a trade-off that exists between adversarial robustness and regularization effect over clean data. Referring to the toy example described earlier, the singular value analysis performed there also supports our claim that methods which impart adversarial robustness tend to suppress sensitivity of the model in all the dominant directions; while the proposed approach provides a nice trade-off by retaining those directions which are essential for modeling the variations in the training data. This ensures that our approach results in better regularization performance on clean data while providing comparable robustness on adversarial data.

\textbf{Results on WideResNet models:} Wide Residual Networks are recently proposed deep architectures that generated state of the art results on CIFAR-10 and CIFAR-100 datasets. In this experiment, we use their publicly available implementation and train them from randomly initialized weights using the proposed approach using the parameter settings described in the experiments section. Specifically, the Ours-joint approach described in the previous section is used for training. As data augmentation, we applied flipping and random cropping as done in their native implementation. The results are shown in Table \ref{tab:cifar-results}. For the compared methods, we quote their published accuracy values. It can be observed that the proposed approach results in improved performance compared to both the baseline model and the model regularized with dropout. This demonstrates the generalization ability of our approach to state of art deep models. 
\begin{center}
\captionof{table}{Classification error rates (\%) on CIFAR-10 and CIFAR-100 for WideResNet (WRN) architectures. Our results are reported as average of 5 runs. For comparison we provide the published WRN baseline results. $^{(*)}$ denotes the results obtained by a single run.}
\label{tab:cifar-results}
\resizebox{.45\textwidth}{!}{%
 \begin{tabular}{| c | c | c | c |} 
 \hline
 Model & \#params & CIFAR-10 & CIFAR-100 \\ [0.5ex] 
 \hline
 WRN-28-10 \shortcite{wideresnet} & 36.5M &  4.00 & 19.25\\ 
 \hline
 WRN-28-10 with dropout \shortcite{wideresnet}  & 36.5M &  3.89 & 18.85\\ 
 \hline
 WRN-40-10 with dropout$^{*}$ \shortcite{wideresnet}  & 51.0M &  3.8 & 18.3\\ 
 \hline
 WRN-28-10 - \textbf{Ours} & 36.5M & \textbf{3.62} $\pm$ 0.05 & \textbf{17.1} $\pm$ 0.1 \\
 \hline  
\end{tabular}
}
\end{center}

\textbf{Generalization to non-image signals:}
In this work, we have considered adversarial perturbations in the space of natural images. The existence of adversarial perturbations has been shown to exist in other types of signals that occur in speech recognition (\cite{serdyuk2016invariant}, \cite{carlini_speech}) and language processing tasks \cite{miyato_NLP}. While the focus of the this paper has been images, there exists a natural extension of our framework to the above modalities. Such an extension is trivially possible since end-to-end learning systems such as deep networks are used in speech and language tasks as well. As future work, we propose to extend the current approach to explore robustness aspects of deep networks trained on modalities other than images.

% ----------------------------
% CONCLUSIONS

\section{Summary and Conclusion}\label{sec:conclusion}
While the behavior of CNNs to adversarial data has generated some intrigue in computer vision since the work of \cite{szegedy2013intriguing}, its effects on deeper networks have not been explored well. We observe that adversarial perturbations for hidden layer activations generalize across different samples and we leverage this observation to devise an efficient regularization approach that could be used to train very deep architectures.  Through our experiments and analysis we make the following observations: (1) Contrary to recent methods which are inconclusive about the role of perturbing intermediate layers of a DNN in adversarial training, we have shown that for very deep networks, they play a significant role in providing a strong regularization (2) The aggregated adversarial effect of perturbing intermediate layer activations is much stronger than perturbing only the input (3) Significant improvement in classification accuracy entails capturing more variations in the data distribution while adversarial robustness can be improved by suppressing the unnecessary variations learned by the network \ref{subsec:toy-eg}. By providing an efficient adversarial training approach that could be used to regularize very deep models, we hope that this can inspire more robust network designs in the future. 

\section{Acknowledgements}
The authors thank Avitas Systems, a GE Venture, for sponsoring this work. This research is based upon work supported by the Office of the Director of National Intelligence (ODNI), Intelligence
Advanced Research Projects Activity (IARPA), via IARPA R\&D Contract No. 2014-14071600012. The
views and conclusions contained herein are those of the authors and should not be interpreted as necessarily representing the official policies or endorsements, either expressed or implied, of the ODNI, IARPA, or the U.S. Government. The U.S. Government is authorized to reproduce and distribute reprints for Governmental purposes notwithstanding any copyright annotation thereon.

\bibliographystyle{aaai}
\bibliography{main.bib}

\end{document}